\documentclass[10pt,twocolumn,letterpaper]{article}
\usepackage{iccv}
\definecolor{cvprblue}{rgb}{0.21,0.49,0.74}
\usepackage[pagebackref,breaklinks,colorlinks,allcolors=cvprblue]{hyperref}
\usepackage{multirow}
\usepackage{cuted}
\usepackage{hyperref}

\def\papername{X-Dancer}
\title{\papername: Expressive Music to Human Dance Video Generation }

\author{
    Zeyuan Chen$^{1,2}$ ~~~~
    Hongyi Xu$^{2}$~~~~
    Guoxian Song$^{2}$~~~~
    You Xie$^{2}$~~~~
    Chenxu Zhang$^{2}$~~~~\\
    Xin Chen$^{2}$~~~~
    Chao Wang$^{2}$~~~~
    Di Chang$^{2,3}$~~~~~
    Linjie Luo$^{2}$~~~~~
    \vspace{1.6pt}
    \\
    $^1$\normalsize UC San Diego~~~~$^2$\normalsize ByteDance~~~~
    $^3$\normalsize University of Southern California~~~~
}
\begin{document}
\maketitle
\vspace{-10cm}

\begin{strip}
    \centering
    \vspace{-6pt}
    \includegraphics[width=0.96\textwidth]{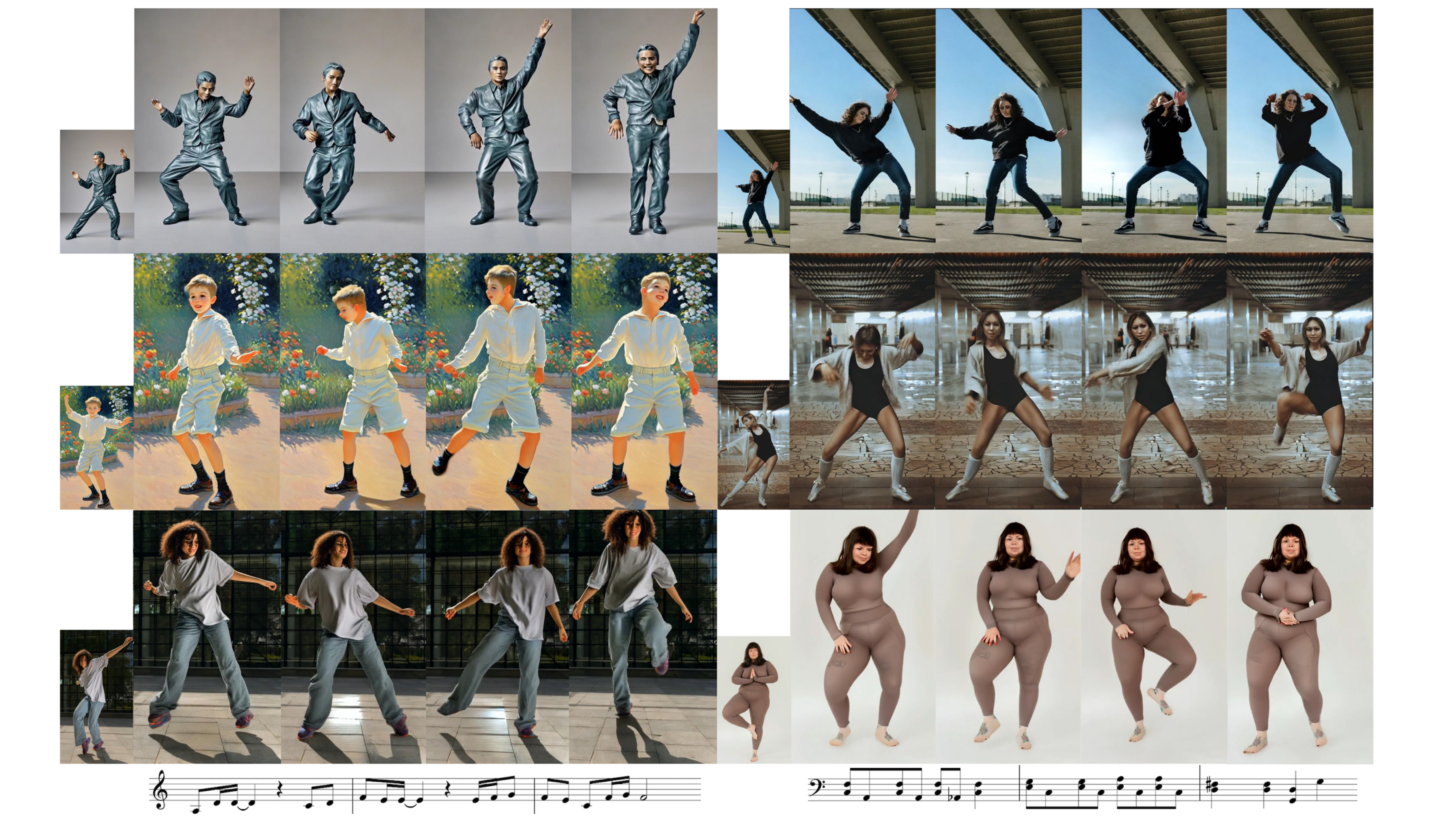}
    \captionof{figure}{
    We present~\papername, a unified transformer-diffusion framework for zero-shot, music-driven human image animation from a single static image. Our approach accommodates diverse body forms and appearances, generating diverse and highly expressive synchronized full-body dance motions. It captures both fine-grained movements of the head and hands and large-scale actions such as body rotations and jumps, seamlessly translating them into vivid and lifelike dance videos.
    }
    \label{fig:teaser}
    \vspace{-2mm}
\end{strip}

\begin{abstract}
We present \papername, a novel zero-shot music-driven image animation pipeline that creates diverse and long-range lifelike human dance videos from a single static image. As its core, we introduce a unified transformer-diffusion framework, featuring an autoregressive transformer model that synthesize extended and music-synchronized token sequences for 2D body, head and hands poses, which then guide a diffusion model to produce coherent and realistic dance video frames. Unlike traditional methods that primarily generate human motion in 3D,  \papername~addresses data limitations and enhances scalability by modeling a wide spectrum of 2D dance motions, capturing their nuanced alignment with musical beats through readily available monocular videos. To achieve this, we first build a spatially compositional token representation from 2D human pose labels associated with keypoint confidences, encoding both large articulated body movements (e.g., upper and lower body) and fine-grained motions (e.g., head and hands). We then design a music-to-motion transformer model that autoregressively generates music-aligned dance pose token sequences, incorporating global attention to both musical style and prior motion context.  Finally we leverage a diffusion backbone to animate the reference image 
with these synthesized pose tokens through AdaIN, forming a fully differentiable end-to-end framework.
Experimental results demonstrate that \papername~is able to produce both diverse and characterized dance videos, substantially outperforming state-of-the-art methods in term of diversity, expressiveness and realism. See our project page for more results: \url{zeyuan-chen.com/X-Dancer/}
\end{abstract}    
\section{Introduction}
\label{sec:intro}

Dance is a universal form of self-expression and social communication, deeply embedded in human behavior and culture. With the rise of social media platforms like TikTok and YouTube Shorts, people increasingly share self-expressive dance videos online. However, creating expressive choreography typically demands practice and even professional training. From a computational aspect, generating realistic dance movements is challenging due to the freeform, personalized nature of dance and its alignment with musical rhythm and structure. In this work, we tackle the challenge of creating continuous, expressive and lifelike dance videos from a single static image, driven solely by a music track.

This study addresses two key challenges in music-driven human image animation: (1) generating smooth, diverse full-body movements at finer scales that capture the complex, non-linear synchronization with music inputs, and (2) translating these generated body movements into high-fidelity video outputs that maintain visual consistency with the reference image and ensure temporal smoothness. Prior approaches~\cite{bailando,edge} mainly focus on computationally generating 3D human poses, such as SMPL skeletons~\cite{loper2023smpl}, from music inputs, utilizing diffusion- or GPT-based frameworks. While these methods excel in producing high-quality, clean motions, they are constrained by limited training datasets (primarily the multi-view AIST dataset~\cite{aist-dance-db,aist++}), which lack diversity, and contain 3D body poses only (excluding head and hands motions). Expanding these datasets with widely available 2D monocular dance videos would require 3D human pose estimation, which is often error-prone and risks degraded motion quality and consistency. Moreover, we target at photorealistic dance video generation rather than 3D skeleton or mesh animations. 

With recent advances in diffusion models,  numerous works have leveraged their generative capabilities to synthesize visually compelling videos by animating a reference image with motion signals such as 2D skeletons~\cite{hu2024animate,chang2023magicpose,hu2023animateanyone}, dense poses~\cite{xu2024magicanimate}, and depth maps~\cite{zhu2024champ}. Unlike motion transfer settings that derive motion from a driving video, our goal is to generate motion signals that are consistent with the reference body shape and aligned with the input musical beats. Recently, a few studies~\cite{hallo,tian2024emo} have attempted to synthesize visual outputs end-to-end from audio inputs using diffusion models. While these methods have advanced in realism and dynamic quality, they still struggle to capture long-range motion and audio context due to high computational demands. Moreover, these frameworks have primarily been validated on audio-driven talking heads, leaving uncertain whether they can accommodate the complexities of full-body kinematics and rapid motion transitions.

To this end, we propose \papername, a unified framework that integrates an autoregressive transformer model for generating extended dance sequences attuned to input music, coupled with a diffusion model to produce high-resolution, lifelike videos. In contrast to prior methods focused on 3D music-to-dance motion generation, our approach models 2D human motion, leveraging widely accessible dance videos where 2D pose estimation is both reliable and scalable. For effective autoregressive motion generation, we develop a multi-part tokenization scheme for per-frame 2D whole-body poses, incorporating detected keypoint confidences to capture multi-scale human motions with motion blur and various occlusions. Thereafter we train a cross-modal transformer that auto-regressively predicts future N-frame pose tokens, paired with per-frame music features extracted with Jukebox~\cite{dhariwal2020jukebox} and Librosa~\cite{jin2017towards}. Our design enables the model to capture a broader diversity of expressive, music-aligned movements with enhanced scalability in both model complexity and data scale. We then leverage a T2I diffusion model with temporal layers to animate the reference image by implicitly translating the generated confidence-aware motion tokens into spatial guidance via a trainable motion token decoder. Specifically, it integrates motion tokens through AdaIN~\cite{huang2017arbitrary} along upsampling from a learnable feature map into multi-scale spatial feature guidance.
By co-learning pose translation with temporal and appearance reference modules~\cite{guo2023animatediff,cao2023masactrl}, the diffusion backbone interprets motion tokens with temporal context and body shape reference,  leading to better shape-disentangled pose control,  smoother and more robust visual outputs even under low-confidence or jittering pose sequences.  
This design establishes an end-to-end transformer-diffusion learning framework, merging the transformer’s strengths in long-context understanding and generation with the diffusion model’s capability in high-quality visual synthesis.

To the best of our knowledge, \papername~represents the first music-driven human image animation framework.
Trained on a large curated music-dance video dataset, our method excels at generating diverse, expressive and detailed whole-body dance videos attuned to input music, adaptable to both realistic and stylized human images across various body types. We extensively evaluate our model on challenging benchmarks, and~\papername~outperforms state-of-the-art 
baselines both quantitatively and qualitatively. Additionally, we highlight its scalability and customization capabilities, showcasing scalability across varying model and data scales, as well as fine-tuning to characterized choreography.  We summarize our contributions as follows,
\begin{itemize}[nolistsep,leftmargin=*]
\item A novel transformer-diffusion based music-to-dance human image animation, achieving state-of-the-art performance in terms of motion diversity, expressiveness, music alignment and video quality.

\item A cross-modal transformer model that captures long-range synchronized dance movements with music features, employing  a multi-scale tokenization scheme of whole-body 2D human poses with keypoint confidence. 

\item A diffusion-based human image animation model that interprets temporal pose tokens and translates them into consistent high-resolution video outputs.

\item Demonstration of captivating zero-shot music-driven human image animations, along with characterized choreography generation.  
\end{itemize}

\section{Related Work}

\subsection{Music to Dance Generation}
Significant progress has been made in realistic human motion generation~\cite{zhi2023livelyspeaker,bailando,edge,lucas2022posegpt,chen2023humanmac,zhang2023generating} from various inputs, such as  speech~\cite{ginosar2019learning,qian2021speech,liu2022learning,zhang2024dr2,ao2022rhythmic} or text~\cite{tevet2023human,jiang2023motiongpt,petrovich2022temos,zhang2024motiondiffuse}. However, the task of music-to-dance generation~\cite{bailando,alexanderson2023listen, edge,li2024lodge,li2024exploring,le2023music,Luo_2024_CVPR,zhang2024bidirectional,wu2021music,zhuang2022music2dance, dabfusion} presents unique challenges: (1) ensuring the generated dance rhythmically aligns with the music, and (2) producing intricate motions with diverse styles and speeds. There has been several 3D human pose datasets~\cite{aist++,li2023finedance,le2023music,Luo_2024_CVPR} proposed. The AIST++ dataset~\cite{aist++} is a 3D music pose dataset containing 1,408 dance motions tailored to various music genres. FineDance~\cite{li2023finedance}, a 3D dataset focusing on fine-grained hand motions, includes 14 hours of dance data. Models like Bailando~\cite{bailando} and EDGE~\cite{edge} have leveraged AIST++ for training. Bailando~\cite{bailando} pioneered using a VQ-VAE for 3D pose encoding, followed by an Actor-Critic GPT model to generate body poses conditioned on music. EDGE~\cite{edge} applied a diffusion framework to predict human poses from a noisy sequence, conditioning on music and employing long-form sampling for extended dance generation. DabFusion~\cite{dabfusion} explores image-to-video generation conditioned on music.
Despite these advancements, existing approaches trained on 3D datasets often generate dance poses with limited diversity. X-Dancer overcomes these data constraints by leveraging a broad spectrum of 2D dance motions, aligning complex movements with musical beats, and utilizing widely available video data to enhance scalability.

\subsection{Human Image Animation}
With the advancements in diffusion models~\cite{ho2022video,guo2023animatediff,blattmann2023stable,BDM,dolfin}, generating high-quality human videos has become feasible. The introduction of ControlNet~\cite{zhang2023adding,omnicontrolnet} and PoseGuider~\cite{hu2024animate,hu2023animateanyone} has further empowered prior methods~\cite{hu2024animate,chang2023magicpose,xu2024magicanimate,zhu2024champ,chang2025x} to create realistic dance videos from a reference image using motion signals such as 2D skeletons, dense poses, and depth maps. 
However, generating expressive, temporally smooth 2D motion sequences that maintain consistency with reference body shapes and align well with music remains an open challenge. 
Several methods~\cite{hallo,jiang2024loopy,tian2024emo} adopt an end-to-end pipeline for generating realistic talking head videos from audio by injecting audio features directly into the network via cross-attention layers. While these methods excel in capturing micro-expressions and lip movements, they fall short in handling highly dynamic and articulated dance movements. X-Dancer implicitly incorporates generated motion tokens into a diffusion backbone to animate the reference image, enabling high-quality and shape-consistent dance video generation.

\begin{figure*}[bt!]
\vspace{-2mm}
\centering
\vspace{-2mm}
\includegraphics[width=0.96\textwidth]{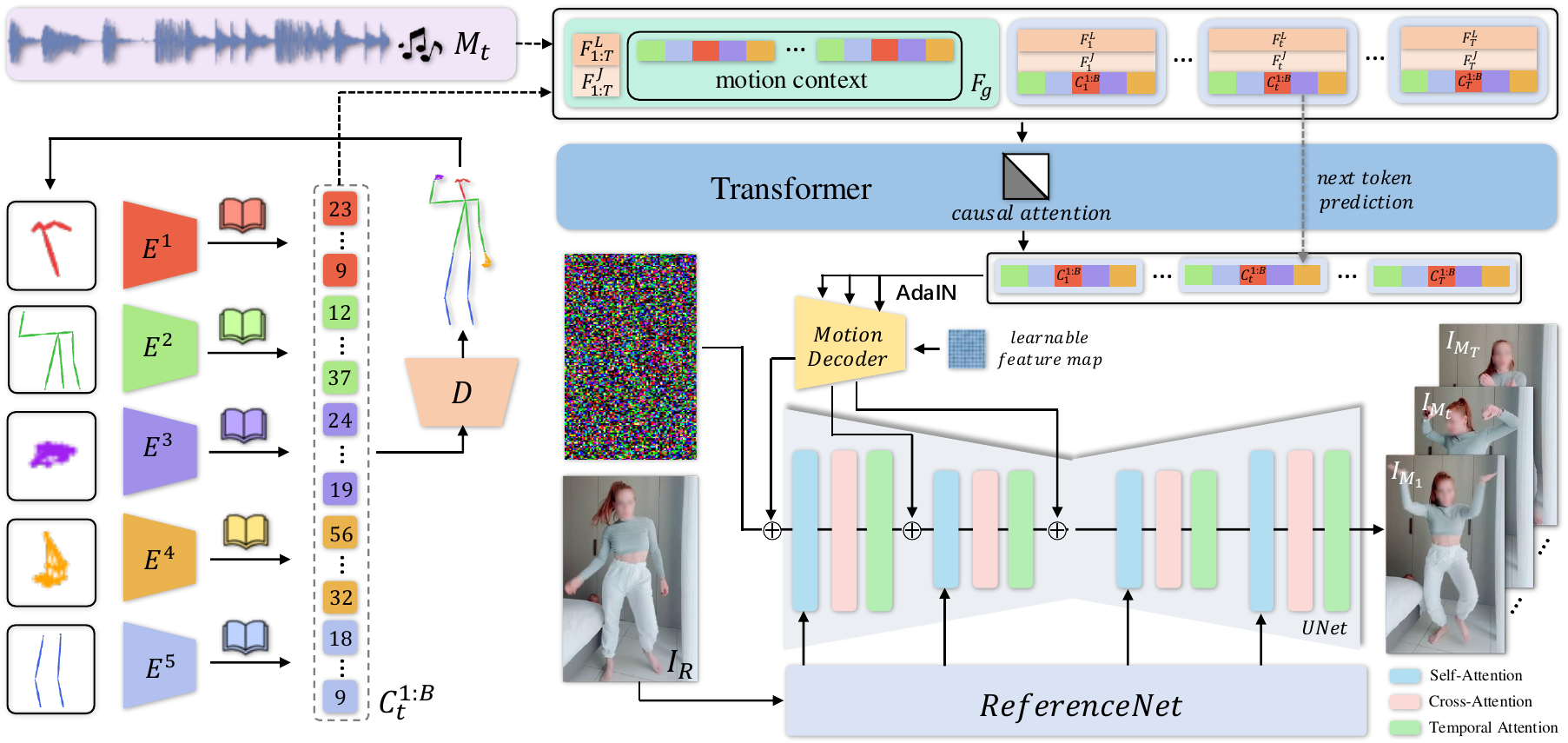}
\vspace{-2mm}
\caption{\textbf{Overview of~\papername.} 
We propose a cross-conditional transformer model to autoregressively generate 2D human poses synchronized with input music, followed by a diffusion model to produce high-fidelity videos from a single reference image $I_R.$ First, we develop a multi-part compositional tokenization for 2D poses, encoding and quantizing each body part independently with keypoint confidence. A shared decoder merges these tokens into a whole-body, confidence-aware pose. Next, a GPT-based transformer autoregressively predicts future pose tokens with causal attention, conditioned on past poses and aligned music embeddings, as well as global music styles and prior motion context. A motion decoder is trained to generate multi-scale spatial pose guidance upsampled from a learned feature map, integrating the generated motion tokens within a temporal window (16 frames) via AdaIN. By co-training the motion decoder and temporal modules, our diffusion model is capable of synthesizing temporally smooth and high-fidelity video frames, while maintaining consistent appearance with the reference image with a trained reference net. }
\vspace{-3mm}
\label{fig:overview1}
\end{figure*}

\section{Method}
Given a single portrait as the reference image $I_R$ and a conditioning music sequence represented as ${M_t},$ our objective is to generate a sequence of dance frames ${I_{M_t}}$, where $t=1,…, T$ denotes the frame index. The generated sequence ${I_{M_t}}$ seeks to maintain the appearance 
and background context depicted in $I_R$ while present expressive dance movements in harmony with the provided musical rhythm and beats. As illustrated in Fig.~\ref{fig:overview1}, our model is trained in two stages: transformer-based 2D human dance motion generation (Section.~\ref{sec:trans}), and diffusion-based video synthesis (Section.~\ref{sec:diff}) from the generated motion sequence. 

\subsection{Data Representations} 
\noindent\textbf{Video Generation.} To generate human dance videos, we employ Latent Diffusion Models~\cite{rombach2022high} that synthesize
samples in the image latent space, facilitated by a pretrained image auto-encoder.  During training, latent features of images are progressively corrupted with Gaussian noise $\epsilon,$ following the Denoising Diffusion Probabilistic Model (DDPM) framework~\cite{ho2020denoising,song2020score}. A UNet-based denoising framework, enhanced with spatial and temporal attention blocks, is then trained to learn the reverse denoising process. 

We apply human-centric cropping to the training videos of half and full-body dances, yielding a unified resolution of $896\times512$. Rather than modeling intricate pixel-wise movements directly on music, we first establish the correlation between music and human dance movement, which subsequently guides the final visual synthesis. 

\noindent\textbf{Motion Modeling.} In contrast to prior methods that generate 3D human motions, we represent diverse dance motions as 2D pose sequences. Compared to its 3D counterparts, 2D human dance motions are widely accessible from large collections of monocular videos, eliminating the need for complex multiview capture setups or labor-intensive 3D animations. Furthermore, the 2D pose detection is significantly more reliable and robust. To enhance the realism and expressiveness, we model not only large body articulations but also finer details on head movements and hand gestures. Notably, we incorporate keypoint confidence into our pose representation, allowing the model to account for motion blur and occlusions.  Each per-frame whole-body pose with $P$ joints is thus represented as $p \in \mathcal{R}^{P\times 3},$ where the last dimension encodes the keypoint confidence. 

\noindent\textbf{Music Embedding.}  Inspired from~\cite{edge}, we utilize the pre-trained Jukebox model~\cite{dhariwal2020jukebox} to extract rich musical features, supplemented by rhythmic information with one-hot encoding of music beats using an audio processing toolbox Librosa~\cite{jin2017towards}.  We resample and synchronize these extracted embeddings—denoted as $F^{J}_{1:T}$ for Jukebox and $F^{L}_{1:T}$ for Librosa—to the video frame rate, ensuring per-frame alignment between the music and visual elements. 

\subsection{Transformer-Based Music to Dance Motion}
\label{sec:trans}

Given a collection of monocular, single-person, music-synchronized dance videos with paired 2D whole-body pose detections, we aim to model the intricate, non-linear correlation between skeletal dance motion and musical features. To achieve this, we first introduce a compositional, confidence-aware VQ-VAE, which captures diverse and nuanced human dance poses across different body parts. Next, we leverage a GPT-like transformer to autoregressively predict future pose tokens, modeling temporal motion token transitions in synchronization with music embeddings.
\\
\noindent\textbf{Compositional Confidence-Aware Human Pose Tokenization.}  Our approach builds on the standard VQ-VAE framework, trained in a self-supervised manner. Given whole-body 2D poses with associated keypoint confidences $p$,  
a 1D convolutional encoder maps $p$ into a latent pose embedding $z_e(p) = E(p),$ which is then quantized by mapping to its nearest representation $z_q(p)$ within a learnable codebook,  and finally the decoder $D$ reconstructs the pose $\hat{p}$ from $z_q(p).$ However, as prior studies~\cite{bailando,yi2023generating,wang2024holistic} suggest, the dependencies between spatial keypoints are complex, and a vanilla VQ-VAE often struggles to capture subtle pose details, such as finger movements and head tilts, due to information loss during quantization and multi-frequency nature of pose variations across different body parts. 

To improve expressive coverage, we train independent 2D pose encoders $E^j$ and learn $B=5$ separate codebooks $Z^j_q$ for upper and lower half bodies, left and right hands, and head respectively,  allowing the model to spatially decompose 2D whole-body pose variations across different frequencies. With pose partition, distinct body-part codes can be flexibly combined, enhancing the range of expressiveness represented within each individual codebook. To capture part-wise spatial correlations and ensure information flow across body parts, we concatenate the quantized pose latents and feed them into a shared decoder. The resulting embedding is then mapped to reconstructed keypoint coordinates through separate projection heads, enabling joint reconstruction while preserving nuanced part-specific details.

We train encoder and decoder simultaneously with the compositional codebooks with the following loss function:
\vspace{-3mm}
\begin{equation}
    \mathcal{L}_{\text{VQ}} = \sum_{j=1}^B \| \hat{p}^j - p^j ||_2 + \beta \sum_{j=1}^B \|\mathrm{sg} \left[z^j_e(p)\right] -  z^j_q(p) \|_2,
\vspace{-2mm}
\end{equation}
where $\mathrm{sg}$ is a stop gradient operation, and the second term is a commitment loss with a trade-off $\beta.$ We utilize exponential moving average (EMA)~\cite{vqvae} to update codebook vectors and remove the embedding loss $\sum_{j=1}^B \|z^j_e(p) - \mathrm{sg} \left[z^j_q(p)\right] \|_2$ for more stable training. 

\vspace{2mm}
\noindent\textbf{Cross-Conditional Autoregressive Motion Modeling.}
After training the compositional quantized multi-part codebooks, 2D human poses detected in our training videos can be represented as sequences of codebook tokens via encoding and quantization. Specifically, each detected pose is mapped to a sequence of corresponding codebook indices, structured as one-hot vectors indicating the nearest codebook entry for each element. We denote this as $C^j_{1:T} = ((c^j_{1, 1}, …, c^j_{K, 1}), \ldots, (c^j_{1, T},\ldots, c^j_{K, T}))$, 
where $K$ is the number of tokens per body part $j$ in each frame.

With this quantized motion representation, we design a temporal autoregressive model to predict the multinomial distributions of possible subsequent poses for each body part, conditioned on both the input music embeddings $F^{J}_{1:T}$ and $F^{L}_{1:T}$ and the preceding motion tokens. Our motion generation transformer is conditioned on the music inputs in two ways. First, we combine the Jukebox and Librosa music embeddings to form starting tokens, which serves as a global music context, such as styles and  genres, that informs the entire motion sequence generation. Second,  inspired by~\cite{bailando}, we concatenate the frame-wise projected music embeddings with the corresponding motion tokens as inputs to the transformer model, ensuring precise synchronization between the motion and music features. This dual conditioning allows our model to produce both globally coherent and locally synchronized dance movements. We use $T=64$ frames for our autoregressive model training. To handle extended motion sequence generation with consistent motion styles and smooth transitions, we additionally incorporate cross-segment motion context into the transformer model. Specifically, we uniformly sample a subset of 8 frames from the previous motion segment as the motion context, and append them after the global music context inputs. We denote the combined context as $F_g.$

Since we model body parts independently, maintaining coherence in the assembled whole-body poses is essential to avoid asynchronous movements (e.g., upper and lower body moving in different directions). To address this, we leverage mutual information across multi-part motions, designing our model with cross-conditioning across body parts. Specifically, we employ a GPT model to estimate the joint distributions of $C^j_{1:T}$ as follows,
\vspace{-3mm}
\begin{gather}
\phi(C^{1:B}_{1:T} | F_g) = \prod_{t=1}^{T}\prod_{j=1}^{B} \prod_{k=1}^{K} \nonumber\\ \phi\bigl (c^j_{k,t} | c^{1:B}_{1:K, <t}, c^{<j}_{1:K, t}, c^{j}_{<k, t}, F^{J}_{\leq t}, F^{L}_{\leq t}, F_g \bigr )
\vspace{-2mm}
\end{gather}

We structure the cross-conditioning between body parts in two ways: (1) the current motion token is conditioned on all preceding motion information from all body parts, ensuring inter-part temporal coherence; (2) by ordering body parts as upper and lower body, head, and hands, we build the hierarchical dependencies from primary components (upper/lower body) to finer, high-frequency movements (head and hands). Since each body part’s pose is represented with a small set of tokens, we empirically observe that causal attention is sufficient to model the next-token distribution within each part. This modeling strategy preserves motion coherence of each body part as a whole, producing expressive and plausible dance movements.  

Our GPT model is optimized through supervised training using a cross-entropy loss on the next-token probabilities. Notably, because our pose tokenization includes associated keypoint confidences, the transformer also learns to model temporal variations in confidence, such as those caused by motion blur and occlusions, enabling it to capture more realistic motion distributions observed in videos.

\begin{figure*}[t!]\vspace{-5pt}
\centering
\vspace{-4mm}
 \includegraphics[width=0.95\linewidth]{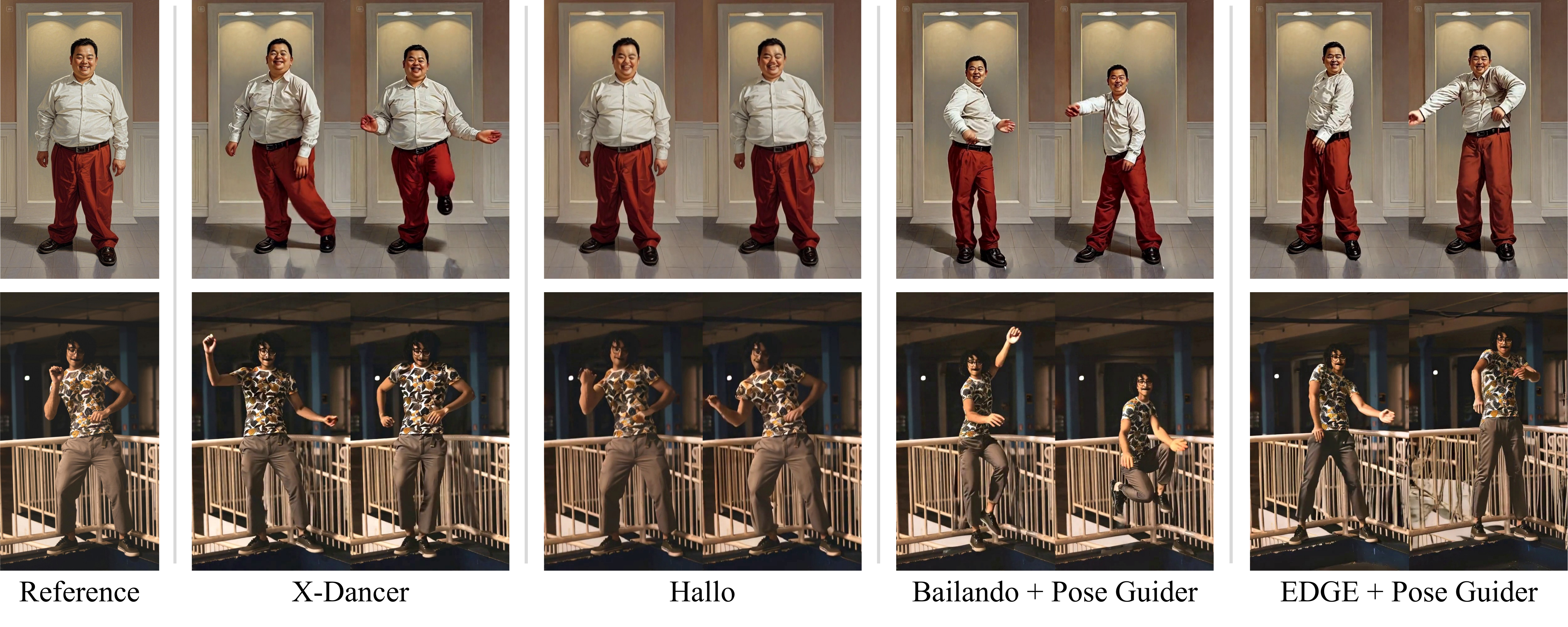}
  \vspace{-6pt}
   \caption{\textbf{Qualitative Comparisons.} Among all the methods, \papername~achieves the most expressive and high-fidelity human dance video synthesis, maintaining the highest consistency with both the reference human characteristics and the background scene.
   }
   \vspace{-6mm}
    \label{fig:baselines}
\end{figure*}

\subsection{Diffusion-Based Dance Video Synthesis}
\label{sec:diff}
We employ a diffusion model to synthesize high-resolution, realistic human dance videos, conditioned on the generated motions from our trained transformer together with a given reference image. To achieve this, we leverage a pretrained T2I diffusion backbone~\cite{rombach2022high} and incorporate additional temporal modules~\cite{guo2023animatediff} for temporal consistency. For transferring the reference image context, a reference network, as a trainable copy of the backbone UNet, extracts reference features of identity appearance and background which are cross-queried by the self-attentions within the backbone UNet. Motion control is achieved through an addition module, often configured in ControlNet~\cite{zhang2023adding} or light-weight PoseGuider~\cite{hu2024animate,hu2023animateanyone}, which translates the motion conditions into 2D spatial guidance additive to the UNet features. 

To incorporate the generated motion tokens into human image animation, one approach is to utilize the trained VQVAE decoder
$D$ to decode pose tokens into keypoint coordinates, which are then visualized as 2D skeleton maps for motion conditioning. These skeleton maps can provide motion guidance to the final diffusion synthesis through a PoseGuider module. While effective to some extent, this approach introduces a non-differentiable process due to the skeleton map visualization, which impedes end-to-end training and often results in the loss of keypoint confidence information.
Additionally, since the temporal module is trained on real video data where pose sequences are typically smooth, it may struggle with jittery or inconsistent motions generated by the transformer model at inference.

In place of explicit translation of motion tokens into 2D skeleton maps as pose guider conditions, we introduce a trainable motion decoder that implicitly and directly translates the 1D pose tokens into 2D spatial guidance, added towards the intermediate features of the denoising UNet. Starting from a learnable 2D feature map, the decoder injects the 1D pose token sequences including the keypoint confidences with AdaIN layers~\cite{huang2017arbitrary}, progressively upsampling this feature map into multiple scales aligned with the resolutions of the denoising UNet features.  The motion decoder is trained alongside the temporal module within a 16-frame sequence window, effectively decoding token sequences into continuous pose guidance that integrates temporal context from adjacent frames. Moreover, by incorporating reference image context during training, we observe empirically that the decoded pose guidance retains minimal identity and body shape information compared to explicit 2D poses, enabling generated pose tokens to adapt seamlessly to subjects with varied body shapes and appearances.
\section{Experiments}

\begin{table*}[t]
    \centering
    \begin{minipage}[t]{0.48\textwidth}
        \centering
        \caption{Quantitative comparison on motion generation.}
        \vspace{-2mm}
        \begin{tabular}{lccc}
            \toprule
            \multirow{2}{*}{\textbf{Metrics}} & \multicolumn{3}{c}{\textbf{AIST++ Dataset / In-House Dataset}} \\
            \cmidrule(lr){2-4}
            & FVD $\downarrow$ & DIV $\uparrow$ & BAS $\uparrow$ \\
            \midrule
            Ground Truth & 509.58/129.75 & 34.10/29.67 & 0.24/0.22 \\
            Hallo~\cite{hallo} & \underline{548.81}/\underline{249.12} & \textbf{28.66}/\textbf{28.98} & 0.16/0.20 \\
            Bailando~\cite{bailando} & 621.22/534.02 & 22.34/24.05 & 0.19/0.19 \\
            EDGE~\cite{edge} & 639.46/303.36 & 24.87/27.29 & \underline{0.26}/\underline{0.24} \\
            \midrule
            \papername-AIST & 620.73/309.18 & 25.25/24.31 & \textbf{0.26}/0.22 \\
            \papername & \textbf{531.52}/\textbf{238.22} & \underline{25.61}/\underline{28.08} & 0.23/0.21 \\
            \bottomrule
        \end{tabular}
        \label{tab:motiongen}
    \end{minipage}%
    \hspace{0.02\textwidth} 
    \begin{minipage}[t]{0.48\textwidth}
        \centering
        \caption{Quantitative comparison on visual synthesis.}
        \vspace{-2mm}
        \begin{tabular}{lccc}
            \toprule
            \multirow{2}{*}{\textbf{Metrics}} & \multicolumn{3}{c}{\textbf{In-House Dataset}} \\
            \cmidrule(lr){2-4}
            & FVD $\downarrow$ & FID-VID $\downarrow$ & ID-SIM $\uparrow$ \\
            \midrule
            Hallo~\cite{hallo} & 609.08 & 76.99 & 0.4870 \\
            Bailando~\cite{bailando} + PG & 583.26 & 100.02 & 0.3392 \\
            EDGE~\cite{edge} + PG & 613.81 & 93.73 & 0.3034\\
            Our motion + PG & 735.05 & 72.71 & 0.4894 \\
            \midrule
            \papername-AIST & 549.38 & 74.06 & 0.3652 \\
            \papername & \textbf{507.06} & \textbf{61.94} & \textbf{0.5317} \\
            \bottomrule
        \end{tabular}
        \label{tab:dancegen}
    \end{minipage}
    \vspace{-4mm}
\end{table*}

\subsection{Implementation Details}
\noindent\textbf{Dataset.} Our model is trained on a curated in-house visual-audio dataset of 76,818 monocular dance recordings in indoor and outdoor settings, averaging 15 seconds per clip. Each video is cropped to $896\times512$ resolution and resampled at 30 fps, covering half- to full-body dances across diverse music and performer characteristics. Details on the dataset are provided in the supplementary paper.

\noindent\textbf{Training.} We train our full pipeline in three stages on 8 A100 GPUs using the AdamW optimizer~\cite{yao2021adahessian}. First, we train a multi-part pose VQ-VAE to encode and quantize 60 joint coordinates and keypoint confidences, into 5-part pose tokens. Each body part's pose uses 6 tokens, each with a unique 512-entry codebook of 6D embeddings. This VQ-VAE is trained for 40k steps with a batch size of 2048 at a learning rate of $2\times10^{-4}$.  Next, we train an autoregressive model for pose token prediction over 300k steps, initialized with pretrained GPT-2 weights, using a batch size of 24 and a learning rate of $1\times10^{-4}$. The model operates on 64-frame pose sequences with a 2224-token context window.
Lastly, we fine-tune the denoising UNet from SD1.5~\cite{rombach2022high} and ReferenceNet with two randomly selected frames per video, followed by co-training the motion token decoder and temporal module~\cite{guo2023animatediff} with diffusion losses on consecutive 16 frames. The diffusion stage is trained for  90k steps at a learning rate of $1\times10^{-5}$ and a batch size of 16. 

\noindent\textbf{Inference.} We initiate our autoregressive dance motion generation from pose tokens encoded from the reference image pose. 
Extended dance sequences are then generated in 64-frame sliding segments with a 12-frame overlap, while 8 uniformly sampled frames from last segment serve as global motion context. We synthesize all video frames simultaneously with the full pose token sequence, applying prompt traveling~\cite{tseng2022edge} to improve temporal smoothness.

\vspace{-2mm}
\subsection{Evaluations and Comparisons}
To the best of our knowledge, no existing approach addresses music-driven 2D video generation from a single human image. We adapted and combined established models to create two baselines for comparisons. First, we adapted the audio-driven, diffusion-based portrait animation model Hallo~\cite{hallo}, retraining it for our task by substituting its audio features with our music embeddings to animate human images via cross-attention modules. For the second baseline, we utilize 3D music-to-dance generation models like Bailando~\cite{bailando} and EDGE~\cite{edge} for motion synthesis, projecting their outputs into 2D pose maps, which are then fed into a diffusion model with a pose guider for controlled image animation. 
For fair comparison, we also train our motion transformer (stage 2, \papername-AIST) on the AIST dataset~\cite{aist-dance-db}, consistent with Bailando and EDGE, but using detected 2D poses instead of 3D skeletons. However, since AIST has limited diversity in identities and appearances, we use our diffusion model—trained on our curated dataset—for comparison on synthesized videos. 
We conduct separate evaluations of all models on the test split of AIST (40 videos) as well as our curated music-video dataset (230 videos).

\begin{table}
    \centering
    \caption{Ablation on our transformer model designs.}
    \vspace{-2mm}
    \begin{tabular}{lccc}
        \toprule
        & FVD $\downarrow$ & DIV $\uparrow$ & BAS $\uparrow$ \\
        \midrule
        w/o global music context & 265.73 & 27.04 & 0.2142 \\
        w/o global motion context & 247.54 & 26.42 & 0.2154 \\
        sub-dataset + GPT-medium & 402.63 & 24.40 & 0.2112 \\
        sub-dataset + GPT-small & 332.93 & 24.58 & 0.2046 \\
        \midrule
        \papername & \textbf{238.22} & \textbf{28.08} & \textbf{0.2182} \\
        \bottomrule
    \end{tabular}
    \vspace{-4mm}
    \label{tab:ablate}
\end{table}

\noindent \textbf{Quantitative Evaluation.} We numerically compare \papername~with baselines in terms of quality of both motion generation and video synthesis. Specifically, we calculate the Fréchet Video Distance (FVD)~\cite{unterthiner2018towards,ge2024content} between generated dance motions and ground-truth training sequences for assessment of motion fidelity. 
For motion diversity, we compute the distributional spread of generated pose features (DIV)~\cite{aist++,bailando},
To numerically evaluate the alignment between the music and generated dancing poses, we follow~\cite{aist++,bailando} to measure the Beat Align Score (BAS) by the average temporal distance between each music beat and its closest dancing beat. These evaluations are conducted in 2D pose space, where we detect poses over synthesized videos from the retrained Hallo model, and 
project the 3D skeleton motion of Bailando and EDGE into 2D via orthographic projection, using camera parameters that best align with the human bounding box in the reference image. 

As shown in Tab.~\ref{tab:motiongen}, our method surpasses all baselines in terms of motion fidelity (FVD) and music beat alignment (BAS), while achieving the second-best diversity (DIV). 
Notably, Bailando and EDGE is trained on professional dances (AIST++~\cite{aist++}), whereas our dataset comprises videos of everyday individuals, reflected in lower BAS even for ground-truth videos. 
Our method trained on AIST (\papername-AIST) achieves the highest BAS whereas our model trained on our curated videos 
still significantly outperforms Bailando and Hallo for beat alignment. 
Hallo entangles motion generation and video synthesis, and achieves a higher DIV score than~\papername~primarily due to its extremely noisy video outputs which results in jittering and chaotic skeleton motions following pose projection.

\begin{table}
    \centering
    \caption{Ablation on VQ-VAE and video diffusion. PG and MD denotes PoseGuider and Motion Decoder respectively. }
    \vspace{-2mm}
    \begin{tabular}{lc}
        \toprule
        & Pose L1 FullBody/Head/Hands $\downarrow$ \\
        \midrule
        Single-Part VQVAE & 0.83 / 0.64 / 0.52 \\
        Multi-Part VQVAE  & \textbf{0.50} / \textbf{0.40} / \textbf{0.42} \\
        \midrule
        & Video PSNR $\uparrow$ / LPSIS $\downarrow$ / FVD $\downarrow$\\   
        \midrule
        GT Pose + PG  & 19.465 / 0.197 / 294.52 \\
        \midrule
        Single-Part + MD  & 17.079 / 0.258 / 350.73 \\
        Multi-Part + PG & 18.836 / 0.207 / 384.16  \\
        \midrule
        Multi-Part + MD & \textbf{19.148} / \textbf{0.207} / \textbf{295.87} \\
        \bottomrule
    \end{tabular}
    \vspace{-4mm}
    \label{tab:ablate_multipart}
\end{table}

For evaluation of video synthesis fidelity, we measure the FVD and FID-VID~\cite{balaji2019conditional} score between the ground-truth and generated dance videos. Additionally, we assess identity preservation using the ArcFace score~\cite{deng2019arcface}, which measures the cosine similarity of face identity features (ID-SIM). All metrics are evaluated on our test video dataset. As an extra baseline, we replace the motion token decoder in our full pipeline with a pose guider (Our motion + PG). As shown in Tab.~\ref{tab:dancegen}, our method achieves the highest visual quality and identity preservation, which we attribute to our disentangled design of motion generation and video synthesis (compared to Hallo) and the use of an implicit motion token decoder rather than an explicit pose guider.

\noindent \textbf{Qualitative Comparisons.}
We present visual comparisons between our method and the baselines in Fig.~\ref{fig:baselines}. For more dynamic comparisons and user study, please refer to our supplementary material. The modified Hallo~\cite{hallo} represents an end-to-end diffusion pipeline that directly synthesizes the final video without intermediate motion generation. However, it exhibits noticeable artifacts, particularly in large articulated body motions, and often fails to preserve the human body's articulation topology.  Bilando~\cite{bailando} and EDGE~\cite{edge} generate body motion in 3D space without considering the scene context or the human shape in the reference image. Despite post-processing for skeleton alignment in the pose guider input, these methods still struggle with significant identity and shape distortions, often producing unnatural interactions with the background scene. Furthermore, they do not model head and hand motions, leading to more rigid and less expressive dance movements compared to~\papername. We also note that our model trained on monocular dance videos is efficient and versatile, easily adapting to specific dances. This is not easily achievable with EDGE or Bailando, which require extensive effort in crafting 3D skeleton dances (supplementary paper Sec.3)

\subsection{Ablation Study}
We conduct ablation studies by systematically removing individual design components from our full training pipeline. 

\begin{figure}[t]
\centering
\includegraphics[width=1.0\linewidth]{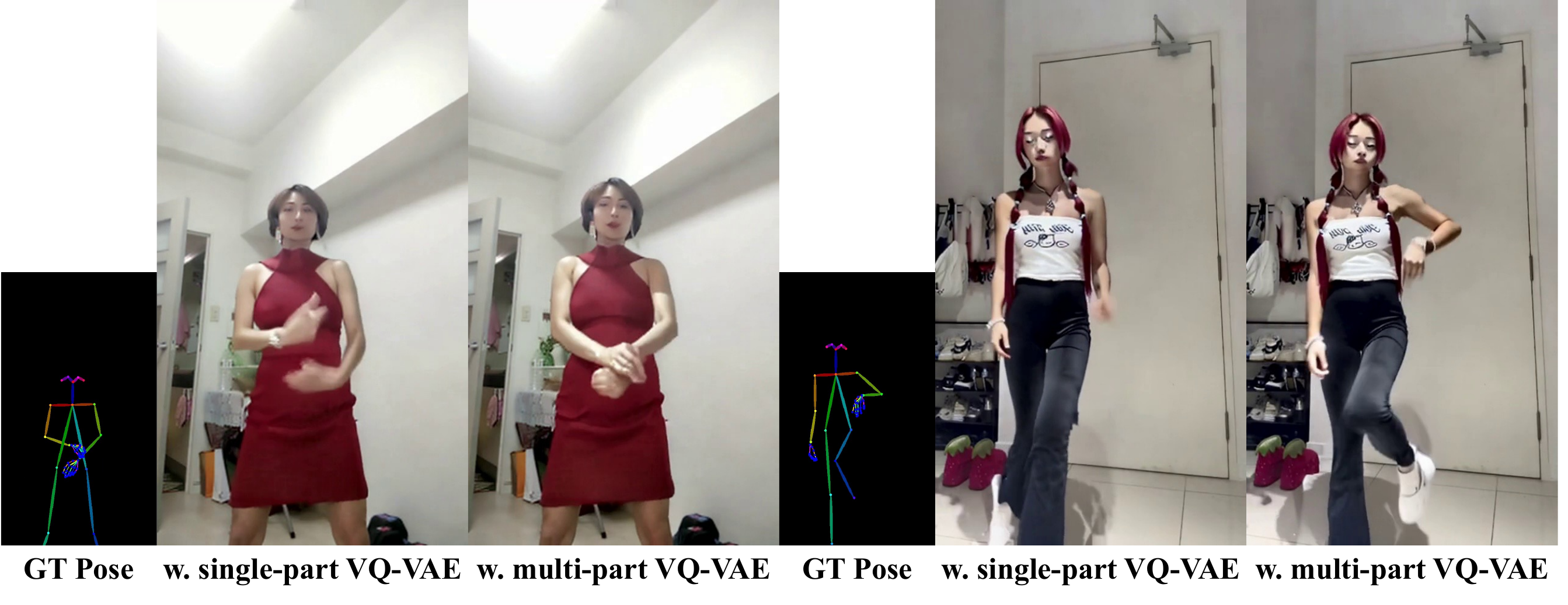}
\vspace{-4mm}
\caption{Our Multi-part VQ-VAE  captures intricate poses such as hand gesture (left), arm bending, head tilting and leg lifting(right) whereas single-part pose tokenization fails to fully replicate them. }
\vspace{-4mm}
\label{fig:multivssingle}
\end{figure}

\noindent \textbf{Multi-part VQ-VAE.} 
As shown in Tab.~\ref{tab:ablate_multipart}, our compositional multi-part VQ-VAE significantly reduces pose reconstruction error compared to single-part whole-body pose tokenization. The multi-part VQ-VAE better captures fine-scale pose variations, while single-part tokenization often loses high-frequency details, resulting in more rigid and less expressive motion generation. To further assess its impact, we train a video generation diffusion model using single-part pose tokenization. As illustrated in Figure~\ref{fig:multivssingle}, the single-part representation struggles to control fine-grained motions like hip swaying, leg lifting and head tilting, leading to perceptually lower-quality videos. This is further supported by its inferior video reconstruction metrics (PSNR and LPIPS~\cite{zhang2018perceptual}) in Tab.~\ref{tab:ablate_multipart}. 

\noindent \textbf{Motion Generation.} 
Next, we assess the impact of global music and motion context on motion generation. As presented in Tab.~\ref{tab:ablate}, both contexts contribute to producing more consistent, plausible, and music-synchronized motions. We further analyze the benefits of 2D motion modeling and transformer-based autoregressive generation by scaling both model parameters and dataset size. Across all metrics, we observe significant performance gains (Tab.~\ref{tab:ablate}) as the number of training parameters increases from 117M (GPT-small) to 345M (GPT-medium) and data scale from 10k to 100k videos, underscoring the scalability potential of monocular dance video data with our architecture, sheding light on further performance improvements as they scale. 

\noindent \textbf{Pose-Guided Video Synthesis.}
In Tab.~\ref{tab:dancegen}, we compare our motion token decoder (X-Dancer) against a pose guider using explicitly decoded skeleton map (Our motion + PG). While our transformer-generated motion exhibits jittering, our motion token decoder significantly reduces jitter and enhances temporal consistency by leveraging temporal motion context. Additionally, it demonstrates superior identity and body shape preservation compared to the pose guider.

In Tab.\ref{tab:ablate_multipart}, we evaluate the effectiveness of our motion token decoder (Multi-Part + MD) against the pose guider (Multi-Part + PG) in self-driven video synthesis. Our motion decoder directly infers body pose coordinates and confidences from pose tokens, achieving lower reconstruction errors (PSNR and LPIPS) compared to the pose guider, which relies on explicitly decoded pose maps. However, due to motion blurriness in fast dance motions and fine-grained structural variations in small pixel regions like the face and hands, visual rendering artifacts remain present in both methods. Notably, these artifacts stem primarily from the diffusion model’s limitations rather than errors in pose reconstruction, as evidenced by the similar video error observed in GT Pose+PG (Tab.~\ref{tab:ablate_multipart}).
\section{Conclusion}

We present \papername, a novel framework that unites an autoregressive transformer with a diffusion model to generate high-quality, music-driven human dance videos from a single reference image. Unlike prior works, \papername\ models and generates dance movements in 2D space, harnessing widely accessible 2D poses from monocular dance videos to capture diverse, expressive whole-body motions. Our method achieves state-of-the-art results in video quality, motion diversity and expressiveness, providing a scalable and adaptable solution for creating vivid, music-aligned dance videos across various human forms and styles.

\noindent\textbf{Limitations and Future Work.} Our model is trained solely on curated real-human daily dance videos, which can be noisy and lack the motion precision found in professional dancer videos.  Consequently, out-of-domain human images may lead to rendering artifacts, and the generated dance motions may occasionally lack music alignment. More failure cases are present in the supplementary paper. While we designed our pipeline to be end-to-end trainable and scalable, we currently implement it in stages due to memory limitations. Future work will explore large-scale, multi-machine training to further enhance performance and efficiency.

\noindent\textbf{Ethics Statement.} Our work aims to improve human image animation from a technical perspective and is not intended for malicious use like fake videos. Therefore, synthesized videos should clearly indicate their artificial nature.

{
    \small
    \bibliographystyle{ieeenat_fullname}
    \bibliography{main}
}
\clearpage
\setcounter{page}{1}
\maketitlesupplementary

In the supplementary material, we provide additional details on our curated in-house dataset (Section~\ref{sec:dataset-supp}). In Section~\ref{sec:userstudy}, we provide a user study to further compare~\papername\ to all baselines. We discuss more use cases of ~\papername\ in Section~\ref{sec:vis}, and present some failure cases and discuss some limitations of our method (Section~\ref{sec:limitation}).  For more dynamic visual results, please refer to our offline webpage.

\section {Dataset}
\label{sec:dataset-supp}
Our in-house video dataset is sourced from a third-party curation service, featuring monocular recordings of everyday dance performances by diverse individuals worldwide. To ensure quality, we filter out videos shorter than 5 seconds, those with low resolution or poor quality (HyperIQA score~\cite{su2020blindly} below 40), videos containing multiple people or with a human bounding box smaller than $27.5\%$ of the frame, and footage with non-static cameras (corner pixel standard deviation exceeding 20). After filtering, the dataset comprises 76,818 dance videos, totaling around 360 hours. All filtered videos are center-cropped to $896\times512$ resolution and resampled to 30 fps.

To analyze the dataset, we employ the state-of-the-art vision-language model (VLM) Qwen2.5-VL~\cite{Qwen2.5-VL}, yielding the following statistics:
\begin{itemize}
    \item Gender distribution: $78\%$ female, $22\%$ male.
    \item Age distribution: $68\%$ of dancers are classified as young, with the rest as middle-aged or older.
    \item Ethnicity distribution: 49\% White, 26\% Asian, 11.2\% Latino, 9\% Black, with the rest others.
    \item Video duration distribution: 10.04\% of videos are between 5-10 seconds, 33.06\% between 10-15 seconds, 49.72\% between 15-20 seconds, 6.25\% between 20-25 seconds, and 0.94\% exceed 25 seconds.
    \item Recording environment: 88\% of videos are recorded indoors.
    \item Dance styles: 89\% are categorized as freestyle/hip-hop, while the remaining consist of popping and locking.
    \item Motion characteristics: 52\% feature medium-level movements (e.g., hip swaying, arm waving), while 47.4\% exhibit strong motion variations (e.g., leg lifts, body rotations, and translations). 96\% of videos do not contain large body turning. 
\end{itemize}
Different from the AIST~\cite{aist-dance-db} and FineDance~\cite{li2023finedance} dataset which capture professional dancers in a multiview setup, our dataset consists of monocular recordings of everyday individuals, offering wider accessibility and greater diversity in both dance motions and identity features. However, compared to professional dancers, our curated performances may exhibit weaker beat alignment, mostly frontal movements and less distinct genre characteristics, often leaning towards freestyle movement. 

\section{User Study}
\label{sec:userstudy}
In addition to our quantitative and qualitative comparisons to various baseline methods provided in Section 4.2 of our main paper, including Hallo~\cite{hallo}, Bailando~\cite{bailando} + PoseGudier, and EDGE~\cite{edge} + PoseGuider. We conduct a user study to further assess perceptual quality, motion fidelity and consistency with the music across all methods. 

We generated 25 dance videos for each method conditioned on randomly selected reference images with music tracks. Each video is 10 seconds long, recorded at 30 FPS, totaling 300 frames per video. Evaluation was conducted using a questionnaire distributed via Google Sheets. Participants were asked to choose the best-performing video among candidates based on five specific metrics: (1) Human Identity Consistency, (2) Music Beat Alignment, (3) Music Style Alignment, (4) Motion Consistency and Naturalness, and (5) Overall Quality. In total, 375 responses were gathered from 15 independent participants with no prior knowledge of video generation techniques. For each question sample, the videos from different methods were randomly permuted. The reference images were also provided to assist participants in assessing identity preservation.

Our method substantially surpasses all the baselines across different metrics, as evidenced in Table~\ref{tab:user}. 

\begin{table*}[h]
    \centering
    \caption{User study. The values indicate user preference ratios (\%).}
    \vspace{-1mm}
    \begin{tabular}{lccccc}
        \toprule
        \textbf{Method} & \textbf{ID Consistency} & \textbf{Beat Alignment} & \textbf{Style Alignment} & \textbf{Motion Consistency} & \textbf{Overall} \\
        \midrule
        Hallo~\cite{hallo} & 18.67 & 2.13 & 1.60 & 1.87 & 1.87 \\
        Bailando~\cite{bailando}+PG &  3.73 & 10.93 & 6.40 & 3.73 & 3.47 \\
        EDGE~\cite{edge} + PG &  4.80 & 24.80 & 17.87 & 11.47 & 9.87 \\
        \midrule
        \papername & \textbf{72.80} & \textbf{52.13} & \textbf{74.13} & \textbf{82.93} & \textbf{84.80} \\
        \bottomrule
    \end{tabular}
    \vspace{1mm}
    \label{tab:user}
\end{table*}

\section{More Visual Results.}
\label{sec:vis}
Please refer to our webpage for all the dynamic visual results. 

\begin{figure}[t]
\centering
\includegraphics[width=0.98\linewidth]{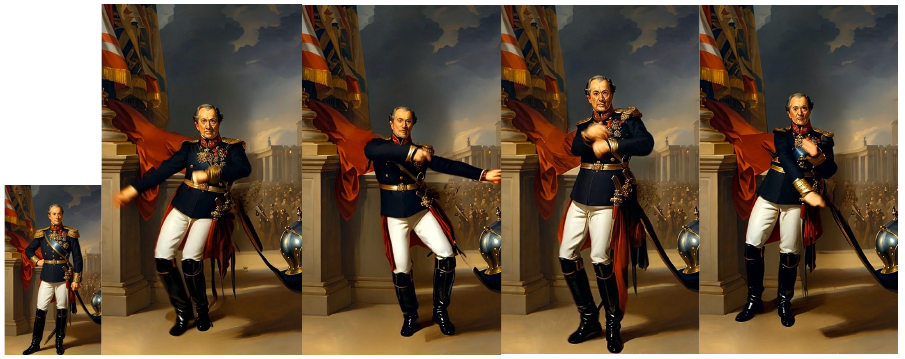}
\vspace{-2mm}
\caption{Human image animation after finetuning motion transformer with 30 dance videos of Subject Three. }
\vspace{-5mm}
\label{fig:characterized}
\end{figure}

\noindent\textbf{Single Reference, Multiple Music.} 
Given a single reference image, \papername~demonstrates the ability to generate diverse and expressive dance motions, maintaining consistent movement styles that adapt to different music genres and beat flows. This underscores \papername's capability to effectively interpret the global music context while synchronizing seamlessly with local rhythmic beats.

\noindent\textbf{Single Music, Multiple References.} 
We present diverse dance videos generated by \papername~for various reference images, all driven by the same music track. While maintaining a consistent dance style synchronized to the shared music, each generated video also reflects the personalized attributes derived from its corresponding reference image, showcasing \papername's adaptability and attention to individual identity details.

\noindent\textbf{Single Music, Single Reference.} 
We demonstrate a variety of dance movements generated from a single reference image, all driven by the same music track. While all dance movements are well-aligned with the music beats, our model exhibits the ability to produce diverse and dynamic dance motions, highlighting its versatility and creativity.

\noindent \textbf{Finetuning for Characterized Choreography.} While our method operates as a zero-shot pipeline, generalizing seamlessly to new reference images and music inputs, it can also be fine-tuned for characterized choreography using only a few sample dance videos. This adaptability is challenging for 3D motion generation models like EDGE~\cite{edge} and Bailando~\cite{bailando}, which require intricate multiview captures or extensive effort in creating 3D dance movements. As shown in Fig.~\ref{fig:characterized} and our supplementary video, our method successfully captures and mimics the specific choreography after fine-tuning with only 30 dance videos from diverse performers, showcasing its efficiency and versatility in adapting to specific dance styles. 

\noindent\textbf{Additional Results.} 
We provide additional results including baseline comparisons and results of \papername-AIST (discussed in line 449 - 455 of the main paper).

\section{Limitations and Failure Cases.}
\label{sec:limitation}

\begin{figure}[h]
\centering
\includegraphics[width=0.98\linewidth]{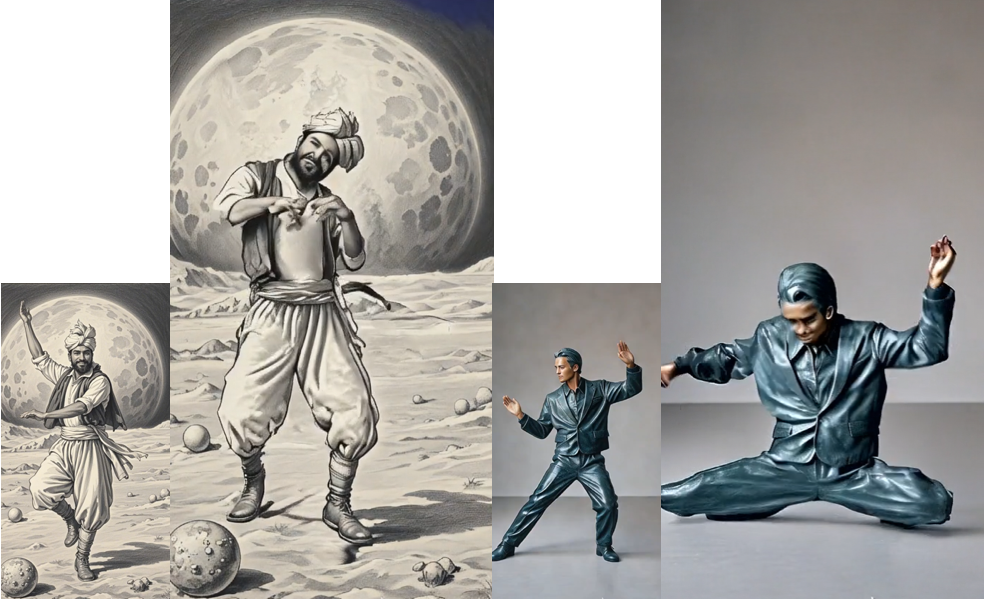}
\vspace{-2mm}
\caption{Failure cases. Please note the rendering artifacts on the face and hands, as well as the unnatural renderings under challenging poses. }
\vspace{-5mm}
\label{fig:artifacts}
\end{figure}

In addition to the limitations and future work discussed in Section 5 of our main paper, we would like to discuss some additional limitations and failure cases. Specifically, noticeable rendering artifacts remain observable, particularly in the face and hands, and color flickering or over-saturation can occasionally occur. However, as indicated in Table 4 of the main paper (GT Pose + PG), these artifacts persist even when using ground-truth poses, and the overall video quality improvement remains marginal. Such issues are challenges across human image animation models and are orthogonal to the core contributions of our work . Future work could explore stronger video diffusion base models such as~\cite{sora,yang2024cogvideox,kong2024hunyuanvideo}, and incorporate specialized embeddings or priors for these fine-grained regions~\cite{narasimhaswamy2024handiffuser} to mitigate these problems.
Furthermore, while our method effectively generates motions informed by the identity and shape of the reference image, there are cases where the generated motions become inconsistent with the background scene or reference identity. This misalignment can lead to unnatural motions and, in severe cases, pronounced rendering artifacts (Figure~\ref{fig:artifacts}).

\end{document}